\documentclass{article}
\usepackage{ijcai22}
\usepackage{times}

\usepackage{soul}
\usepackage{url}
\usepackage[hidelinks]{hyperref}
\usepackage[utf8]{inputenc}
\usepackage[small]{caption}
\usepackage{graphicx}
\usepackage{amsmath}
\usepackage{booktabs}
\usepackage{fixltx2e}
\usepackage{chronology}
\usepackage{comment}
\usepackage[permil]{overpic}
\usepackage{todonotes}

\usepackage{color}

\usepackage{soul} 

\usepackage{enumitem}
\setlist[enumerate]{leftmargin=5mm}

\title{Next-Year Bankruptcy Prediction from Textual Data: \\ Benchmark and Baselines}

\author{
Henri Arno$^1$\footnote{Contact Author} \and Klaas Mulier$^1$ \and Joke Baeck$^1$ \And Thomas Demeester$^2$
\affiliations
$^1$Ghent University\\
$^2$Ghent University - imec\\
\emails
first.last@UGent.be}

\begin{document}

\maketitle

\begin{abstract}
Models for bankruptcy prediction are useful in several real-world scenarios, and multiple research contributions have been devoted to the task, based on structured (numerical) as well as unstructured (textual) data. However, the lack of a common benchmark dataset and evaluation strategy impedes the objective comparison between models. This paper introduces such a benchmark for the unstructured data scenario, based on novel and established datasets, in order to stimulate further research into the task. We describe and evaluate several classical and neural baseline models, and discuss benefits and flaws of different strategies.  In particular, we find that a lightweight bag-of-words model based on static in-domain word representations obtains surprisingly good results, especially when taking textual data from several years into account. These results are critically assessed, and discussed in light of particular aspects of the data and the task.
All code to replicate the data and experimental results will be released.
\end{abstract}

\section{Introduction}
\label{sec:intro}

\begin{table*}[t]
\begin{footnotesize}
\begin{tabular}{ll}
\toprule
\textbf{\begin{tabular}[c]{@{}l@{}}Three years prior \\ to bankruptcy\end{tabular}} & \textit{\begin{tabular}[c]{@{}l@{}}\addlinespace"We are highly leveraged and a substantial portion of our liquidity needs arise from debt service  requirements \\ and from funding our costs of operations and capital expenditures, including  acquisitions... we entered into \\a new asset-based  revolving \ul{credit facility} (ABL Facility)...  secured by substantially all of our assets..." \end{tabular}} \\          \midrule                                                       
\textbf{\begin{tabular}[c]{@{}l@{}}One year prior\\ to bankruptcy\end{tabular}}     & \textit{\begin{tabular}[c]{@{}l@{}} \addlinespace" ... we received a \ul{waiver} of certain events of \ul{default} under the TLA arising  from the inclusion  of a going \\ concern qualification from our registered public accounting firm, breach of the  \ul{EBITDA} \ul{financial covenant}, \\and \ul{cross-default} arising from the \ul{default} under our  ABL Facility... In order to address our liquidity issues and\\ provide for a \ul{restructuring} of our  \ul{indebtedness} to improve our long-term capital structure, we have entered \\ into a \ul{Restructuring}  Support Agreement  ...  pursuant to a prepackaged plan of reorganization to be filed in a case \\ commenced under chapter 11 of the United States Bankruptcy Code..."\end{tabular}}\\
\bottomrule 
\end{tabular}
\caption{\label{tbl:extract}
Extracts from the MD\&A section of a distressed company in our dataset, one year and three years prior to bankruptcy. Underlined words correspond to the top 20 tokens most informative for imminent bankruptcy in our respective Binary Bag-of-Words models. }
\end{footnotesize}
\vspace{-2mm}
\end{table*} 
%

Since the seminal work of \citeauthor{beaver1966financial} [\citeyear{beaver1966financial}], bankruptcy prediction has received considerable attention by both academics and practitioners. A sound prediction model has numerous applications. For instance, successful quantitative methods can help professionals, such as creditors and investors, in managing financial risk \cite{bielecki2013credit}. Furthermore, as \citeauthor{bernanke1981bankruptcy} [\citeyear{bernanke1981bankruptcy}] has shown that economy-wide levels of bankruptcy risk plays a structural role in propagating recession, regulators can use bankruptcy prediction models to monitor the financial health of key economic actors and control systematic risk. 

A large number of bankruptcy prediction models have been proposed in literature, such as the models from \citeauthor{beaver1966financial} [\citeyear{beaver1966financial}], \citeauthor{ohlson1980financial} [\citeyear{ohlson1980financial}], \citeauthor{odom1990neural} [\citeyear{odom1990neural}], \citeauthor{KIM20103373} [\citeyear{KIM20103373}] and \citeauthor{mai2019deep} [\citeyear{mai2019deep}]. However, it appears difficult to compare these studies and objectively assess progress in the field. We have identified the following three aspects that make comparison difficult: (1) the temporal nature and typical class imbalance of the bankruptcy prediction task leads to strongly deviating evaluation scenarios, (2) there is little consensus on the key evaluation metrics, and (3) there is no standard benchmark dataset.  These issues are further discussed in section~\ref{sec:need_benchmark}. In order to overcome these problems, we have designed and described our experimental setup with reproducibility on a common benchmark in mind. To that end, scripts to reconstruct the benchmark and reproduce the presented results are available at \url{https://github.com/henriarnoUG/BankruptcyBenchmarkBaselines}. Note that this paper investigates the potential to predict bankruptcy from textual disclosures only. Extending this benchmark to the hybrid case of combined textual and structured features will be part of our future work. 

The contributions of this paper are as follows: 
(1) we introduce a reproducible benchmark for text-based bankruptcy prediction, based on novel and established economic datasets,
(2) classical as well as neural baseline prediction models are provided, including results on next-year bankruptcy prediction from multiple years of textual data, 
and (3) insights into the results are given along with pointers to potential next steps in bankruptcy prediction.


\section{Related Work}
After a general overview of research on bankruptcy prediction (Section~\ref{subsec:bankruptcy}), we describe some key aspects that make contributions in literature hard to compare (Section~\ref{sec:need_benchmark}).

\subsection{Bankruptcy Prediction Research}\label{subsec:bankruptcy}
\citeauthor{beaver1966financial} [\citeyear{beaver1966financial}] pioneered bankruptcy prediction literature with a discriminant model based on financial ratios. Subsequently, well-chosen structured financial variables were proposed to predict failure, along with increasingly advanced prediction models. 
Statistical models, such as discriminant analysis \cite{beaver1966financial,altman1968financial}, have been dominant in the past but rely on stringent assumptions about the data \cite{balcaen200635}. Today, machine learning models are commonplace as they rely on fewer assumptions and learn directly from the data. \citeauthor{odom1990neural} [\citeyear{odom1990neural}] used neural networks to predict bankruptcy, \citeauthor{KIM20103373} [\citeyear{KIM20103373}] have built an ensemble model and \citeauthor{HOSAKA2019287} [\citeyear{HOSAKA2019287}] generates predictions through a convolutional neural network with ratios presented as images.
%
\citeauthor{keasey1987non} [\citeyear{keasey1987non}] were the first to include non-financial variables in a corporate failure model, \citeauthor{shumway2001forecasting} [\citeyear{shumway2001forecasting}] has shown that market-driven variables are strongly related to bankruptcy and \citeauthor{cecchini2010making} [\citeyear{cecchini2010making}] found that textual disclosures can be used to discriminate between bankrupt and non-bankrupt firms. The information value of textual data was further established by \citeauthor{mayew2015md} [\citeyear{mayew2015md}] as they found that the opinion of management on the future of the company and the linguistic tone of the Management Discussion and Analysis has significant explanatory power for corporate failure. \citeauthor{mai2019deep} [\citeyear{mai2019deep}] provide large-sample evidence of the predictive power of textual disclosures and show that deep learning models yield superior results when using textual data together with traditional accounting features. Furthermore, the authors compare two deep learning architectures based on skip-gram word representations \cite{https://doi.org/10.48550/arxiv.1301.3781} and conclude that an average embedding model leads to better results than a ConvNet architecture. Despite this promising work, bankruptcy prediction models using textual data are scarce.

\subsection{Need for a Reproducible Benchmark}\label{sec:need_benchmark}
The following aspects prevent a straightforward comparison of research contributions, and may be avoided by a common benchmark along with the tools to reproduce experimental results, one of the goals of this work.

\paragraph{Temporal nature and class imbalance of bankruptcy data:}
Due to the temporal nature of the data and the typically much smaller fraction of positive cases (enterprises going bankrupt), many strategies have been proposed to construct training data and define evaluation sets. 
The data source that serves as a basis for the model typically contains annual (or more fine-grained) observations for each firm in the sampling period. In earlier work \cite{beaver1966financial,altman1968financial} the explanatory variables were selected only once for each firm in the dataset. In the `paired sampling' approach \cite{altman1968financial}, the independent variables for failed firms were retained
in the year before failure, together with those for a paired healthy firm in that same year, to induce a balanced dataset from which a random evaluation set is sampled.  
\citeauthor{shumway2001forecasting} [\citeyear{shumway2001forecasting}] has shown that 
such an approach leads to poor out-of-sample prediction performance and incorrect statistical inference.  As an alternative, hazard models can be estimated by treating each firm-year sample as an independent observation, with the bankruptcy status by the end of the following year as the prediction target. Typically, the observations prior to some date are used for model training, and observations after this date are used to estimate the out-of-period prediction performance \cite{shumway2001forecasting,mai2019deep}. Sometimes even a random split is used, independent of time \cite{mai2019deep}.  In the work of \citeauthor{volkov2017incorporating} [\citeyear{volkov2017incorporating}], the explanatory variables for a number of consecutive years are used as input, with company status as the prediction target in the year afterwards. The class imbalance is managed through undersampling of healthy companies. Evaluation is done on a held-out subset of companies, which is therefore artificially balanced as well. Undersampling, oversampling, and data augmentation techniques are investigated by \citeauthor{veganzones2018investigation} [\citeyear{veganzones2018investigation}]. Training and evaluation are done on a non-overlapping subset of firms, with a one-year shift in between, while also maintaining a predefined artificial ratio between the number of healthy and bankrupt firms (for both training and evaluation).

In our considered population (public companies in the US, see Section~\ref{sec:data}), all companies are known, as well as their yearly reports so far, and the goal is predicting bankruptcy for all of these firms in the near future (the coming year).
This is simulated in our evaluation scenario, where we make predictions for \emph{all} companies not (yet) bankrupt and observed through annual reports up to a given year, on their bankruptcy status the year afterwards (as further detailed in section~\ref{sec:dataset_construction}).


\paragraph{Large variety of evaluation metrics:}
The choice of evaluation metrics is often linked to the experimental setup, e.g., depending on whether a balanced test set is used.
The evaluation scenario also influences the choice of threshold used for metrics like accuracy, precision, or recall.  For example, 
\citeauthor{volkov2017incorporating} [\citeyear{volkov2017incorporating}] selects a threshold that maximises the $F_2$-measure.  
Alternatively, \citeauthor{veganzones2018investigation} [\citeyear{veganzones2018investigation}] select the threshold that minimises the expected cost of misclassification with equal weights. Aggregated metrics that avoid the use of a threshold, such as area under the ROC curve (AUC), decile rank, and cumulative accuracy profile ratio (CAP) are regularly reported as well \cite{mai2019deep}.  

\paragraph{Use of private datasets:}
The final reason that makes model comparison hard is the lack of a standard benchmark dataset. Bankruptcy prediction literature either reports results on proprietary datasets \cite{matin2019predicting} or on data obtained by manual collection or custom web scraping strategies (and kept private) \cite{cecchini2010making,wang2020financial}. For a comprehensive overview of data sources used in recent corporate failure literature we refer the reader to the work of \citeauthor{mai2019deep} [\citeyear{mai2019deep}]. Our datasets are based on the combination of existing sources, i.e., the UCLA-LoPucki Bankruptcy Research Database (BRD)\footnote{https://lopucki.law.ucla.edu/} and the public EDGAR-CORPUS \cite{loukas2021edgarcorpus}. This allows researchers to reconstruct the same 
train, validation and test data from these sources, even if we are not allowed to make the resulting datasets public directly.

\section{Methodology}
In the next sections, we describe the data sources (Section~\ref{sec:data}) and motivate our design choices for the benchmark (Section~\ref{sec:dataset_construction}), document pre-processing (Section~\ref{sec:preprocessing}), and the selected evaluation metrics (Section~\ref{sec:metrics}).


\subsection{Data Sources}
\label{sec:data}
Our study makes use of the EDGAR-CORPUS, a novel economic dataset containing 10-k reports from all publicly traded companies in the US, spanning 25 years \cite{loukas2021edgarcorpus}. As we need information on bankruptcies as prediction target, these reports were matched with the UCLA-LoPucki Bankruptcy Research Database (the BRD)\footnote{The BRD does require a paid annual subscription or a one-time purchase for academic single use.}, through the unique Central Index Key to identify companies. The BRD contains information on all Chapter 7 and Chapter 11 filings of the United States Bankruptcy Code since 1997 and is updated monthly. 

Consistent with prior work \cite{cecchini2010making,mayew2015md,mai2019deep}, we limit the 10-k reports to section 7: ``Management Discussion and Analysis''. 
According to the U.S. Securities and Exchange Commission\footnote{https://www.sec.gov/fast-answers/answersreada10khtm.html}, it \textit{``... gives the company’s perspective on the business results of the past financial year. This section, known as the MD\&A for short, allows company management to tell its story in its own words.''} It also contains the risks and uncertainties that could materially affect the company. As an example, consider the extracts from the MD\&A's of a distressed firm in Table~\ref{tbl:extract}.

\begin{figure}[t!]
\hspace{-0.15in}
 \includegraphics[scale = 0.52]{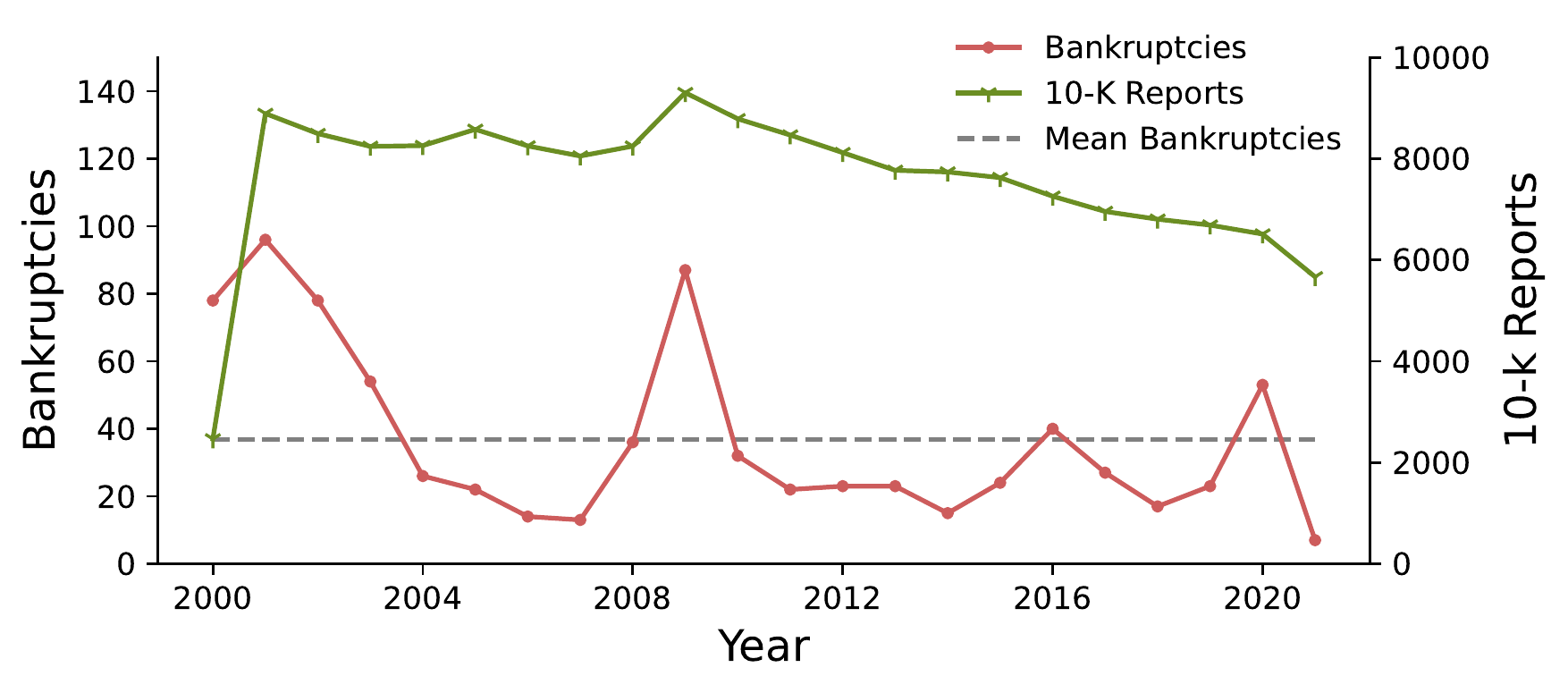}
\vspace{-2mm}
\caption{\label{fig:imbalance}The number of bankruptcies (including the mean) (left y-axis) and the number of 10-k reports filed (right y-axis) per year. }
\end{figure}

Public company bankruptcy is a rare event. 
Figure~\ref{fig:imbalance} shows that the number of 10-k reports filed by non-bankrupt companies heavily exceeds the yearly number of Chapter 7 and Chapter 11 cases. 
Note how the influence of the Dot-com crisis (2000), the financial crisis (2007-2008), and the COVID crisis (2020) on our population can be observed. Table~\ref{tbl:stats} provides additional statistics for the aligned data sources. 

\subsection{Task Definition and Setup}
\label{sec:dataset_construction}
\subsubsection{Determining the prediction time window}
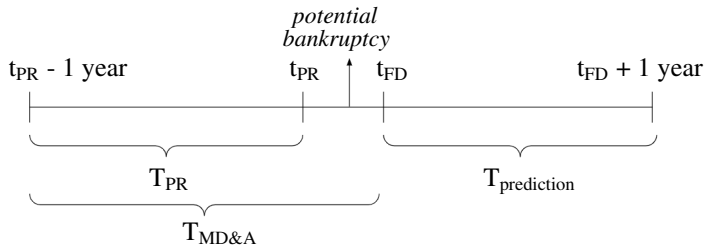
\begin{figure}[t!]
\vspace{0.1in}
\hspace{-0.1in}
\begin{overpic}[width=0.5\textwidth]{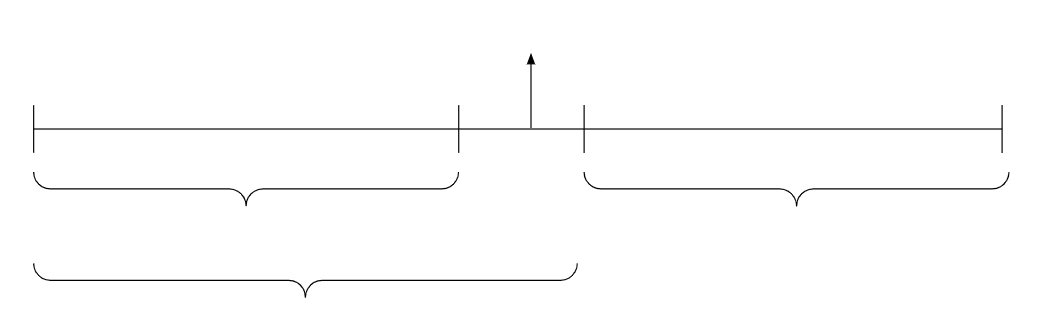}%
\put(420,245){t\textsubscript{PR}}%
\put(550,245){t\textsubscript{FD}}%
\put(210,80){T\textsubscript{PR}}%
\put(260,0){T\textsubscript{MD\&A}}%
\put(0,245){t\textsubscript{PR} - 1 year}%
\put(850,245){t\textsubscript{FD} + 1 year}%
\put(710,80){T\textsubscript{prediction}}%
\put(410,290){\footnotesize\textit{bankruptcy}}
\put(425,325){\footnotesize\textit{potential}}
\end{overpic}
\caption{\label{fig:timeline} Timeline containing the characterising dates (t\textsubscript{PR}, t\textsubscript{FD}) of a 10-k report and corresponding periods (T\textsubscript{PR}, T\textsubscript{MD\&A}, T\textsubscript{prediction})}
\end{figure}

Prior work has not always been very transparent about the temporal aspect of the textual and numerical data in their models, but this requires special attention in order to arrive at a correct setup. 
A 10-k report is characterised by two dates, as schematically shown in Fig.~\ref{fig:timeline}: 
(1) the fiscal year-end t\textsubscript{PR} of the one-year time window T\textsubscript{PR} (`period of report') used to calculate the financial statements, and (2) the filing date t\textsubscript{FD} on which the report is filed with the SEC. Since in practice t\textsubscript{FD} $\geq$ t\textsubscript{PR}, there may be a period after t\textsubscript{PR} yielding textual information in the MD\&A (i.e., before t\textsubscript{FD}), not present in the financial statements. It is therefore important to use the one-year period directly \emph{after} t\textsubscript{FD} as the prediction time window T\textsubscript{prediction} when the textual data is used as input to the model. In the extreme case of bankruptcy in between t\textsubscript{PR} and t\textsubscript{FD} (`potential bankruptcy' in Fig.~\ref{fig:timeline}), it would lead to leakage and artificially high prediction accuracies if the year directly after t\textsubscript{PR} were used for prediction. It is possible, though, that information on an imminent bankruptcy shortly \emph{after} t\textsubscript{FD} is already included in the report, but this does not present a conceptual problem for the prediction setup.

\begin{table}[t!]
\begin{center}
\begin{footnotesize}
\begin{tabular}{@{}ll@{}}
\toprule
period & 2000-2021 \\
avg.~reports per year  & 7599 $\pm$ 1477\\
avg.~bankruptcies per year &  39 $\pm$ 26  \\
avg.~new enterprises per year & 1467 $\pm$ 1376    \\
avg.~doc.~length (\# tokens) &  6492  $\pm$ 1138  \\
\bottomrule
\end{tabular}
\end{footnotesize}
\caption{\label{tbl:stats} Summary statistics of our aligned data sources.}
\end{center}
\vspace{-2mm}
\end{table}

\subsubsection{Dealing with missing 10-k reports}
The dataset contains yearly 10-k reports from the first time a company appears, starting from the year 2000, until 2021 or until bankruptcy.
However, some reports are missing for a number of companies, and our analysis reveals the following three scenarios.
First, some companies stop reporting from a certain point in time onwards, without filing for bankruptcy. This may be due to a merger or an acquisition, but that particular information is not present in the data. 
Second, there may be gaps in the sequence of yearly reports. 
This arises when a company either does not submit a 10-k report (due to unknown reasons) or because of data quality issues. 
Third, we observe that some companies headed towards bankruptcy tend to fail in their reporting in the year(s) leading up to the bankruptcy filing. A naive approach would be to simply discard all instances with missing reports. However, this would make the evaluation scenario biased, since missing reports are not distributed uniformly over the data, due to the different scenarios described above.

Consider our 2019 test set with a history of three years (discussed later in this section) as an example, of which close to 45\% of companies have at least one missing report during the three-year history. The relative frequency of bankruptcy is 0.27\% for the entire population, 0.00\% for companies with only missing data (cf.~an M\&A event), 0.35\% for companies with no missing data and 0.93\% for companies where the data in only the year before prediction is missing. Therefore, we do not remove these companies and keep them in our dataset which results in a more realistic evaluation scenario.

\subsubsection{Construction of input and target per firm-year}
In order to create time-agnostic firm-year samples (following \citeauthor{shumway2001forecasting} [\citeyear{shumway2001forecasting}]) during the construction of our train, validation and test sets (see further),
we process a given year and company as follows:


\begin{enumerate}
    \item \textbf{Determine T\textsubscript{prediction}:} If a 10-k report was filed by the company in the considered year, T\textsubscript{prediction} is the period between t\textsubscript{FD} and t\textsubscript{FD} + 1 year (cf. Figure 2). Otherwise, we use the one-year period starting the same day as the latest available t\textsubscript{FD}, but in the considered year.
    \item \textbf{Assign target label:} If the company filed for bankruptcy during T\textsubscript{prediction}, the label is 1, otherwise 0. Note that potential firm-year instances with a bankruptcy filing \emph{before} t\textsubscript{FD} are invalid for the considered year, as explained above. 
    \item \textbf{Collect textual data:} The MD\&A text from the report filed at t\textsubscript{FD} is used for the one-year history setting, as well as from the two previous years for the three-year scenario. For missing reports, the token `missing' is used.
\end{enumerate}


\subsubsection{Train / validation / test segmentation}
\label{sub:eval}
\paragraph{Training data:} We construct two training sets in total. The first, using data up to 2015, is used for initial training while leaving sufficient data for validation during hyperparameter tuning. 
The second, with data up to 2017, is used to train the final models. They are constructed as follows:
\begin{enumerate}
    \item We leave out all 
    reports with a t\textsubscript{FD} later than 2015 (2017), to ensure a proper temporal split between training and evaluation data. 
    \item For every firm and every year between the first year of the training data and 2015 (2017), we construct a firm-year instance as described above.
    
    
    \item To reduce the impact on the training process of instances without any reports in their considered history (i.e., the one-year or three-year history, respectively), 95\% of those are randomly removed. 
\end{enumerate}

\paragraph{Validation data:}
We construct two validation sets, one for 2017 and one for 2018, both to be used for hyperparameter tuning. First, we filter out companies that have not filed any reports during the 5 years leading up to and including 2017 (2018). For each of these companies, one firm-year sample is created according to the method described above for the year (and hence t\textsubscript{FD}, even when the report is missing) 2017 (2018).


\paragraph{Test data:}
In the same way, we construct two test sets, one for 2019 and one for 2020 (denoting the calendar year containing t\textsubscript{FD}), for the final evaluation of the trained models. 



\subsection{Pre-processing}\label{sec:preprocessing}
When dealing with textual data it is common to perform document pre-processing in order to decrease the dimensionality of the problem and reduce the computational cost of encoding the documents. We perform four pre-processing steps for the Bag-of-Words models presented in sections~\ref{sec:ohe}-\ref{sec:w2v}. First, we lowercase all documents. Second, we remove stopwords and punctuation. Third, we lemmatize each word in the documents through the NLTK library \cite{loper2002nltk}. Inflicted word forms such as \textit{paying} and \textit{payed} are transformed into the root form \textit{pay}. Finally, we replace uncommon words by the token `\_UNK\_' (for `unknown'). A word is deemed uncommon when it does not appear in the 50,000 most frequent words in the training set. When dealing with transformer models \cite{vaswani2017attention}, such as the Longformer \cite{https://doi.org/10.48550/arxiv.2004.05150}, these steps are typically not required and might even lead to deteriorating performance. Preprocessing then consists of proper tokenization of the input text. 
We use the tokenization tools from Huggingface \footnote{https://huggingface.co/}, which allow transforming the input text into a sequence of well-chosen word pieces.

\begin{table*}[t!]
    \centering
\begin{footnotesize}    
\begin{tabular}{@{}lcccc|cccc@{}}
           & \multicolumn{4}{c|}{\textbf{Single year history}}                            & \multicolumn{4}{c}{\textbf{Three year history}}                                     \\ 
           \midrule
           & \multicolumn{1}{l}{Binary} & TF-IDF               & W2V          & Longformer  & \multicolumn{1}{l}{Binary} & TF-IDF               & W2V          & Longformer  \\ \midrule
AUC        & 0.79 (0.84)             & 0.80 (0.85)          & \textbf{0.88 (0.90)} & 0.78 (0.79) & 0.90 (0.92)             & 0.92 (0.96)          & \textbf{0.95 (0.95)} & 0.85 (0.84) \\
AP         & 0.07 (0.05)             & \textbf{0.16 (0.16)} & 0.08 (0.12)          & 0.01 (0.03) & 0.03 (0.06)             & \textbf{0.10 (0.10)} & 0.04 (0.09)          & 0.02 (0.02) \\
recall@100 & 0.19 (0.18)             & 0.26 (0.31)          & \textbf{0.37 (0.31)} & 0.04 (0.07) & 0.15 (0.22)             & \textbf{0.37 (0.29)} & 0.22 (0.24)          & 0.11 (0.02) \\
CAP        & 0.56 (0.68)             & 0.59 (0.72)          & \textbf{0.75 (0.80)} & 0.52 (0.58) & 0.82 (0.84)             & 0.83 (0.92)          & \textbf{0.89 (0.89)} & 0.71 (0.68) \\ \midrule
1          & 0.56 (0.67)             & \textbf{0.74 (0.71)}          & 0.70 (0.73)          & 0.56 (0.51) & 0.78 (0.73)             & 0.70 (0.91)          & \textbf{0.85 (0.84)}          & 0.41 (0.40) \\
2          & 0.74 (0.78)             & 0.74 (0.84)        &   \textbf{0.78 (0.80)}          & 0.70 (0.71) & 0.89 (0.87)             & \textbf{0.93 (1)}             & 0.93 (0.93)          & 0.78 (0.80) \\
3          & 0.78 (0.84)             & 0.78 (0.87)          & \textbf{0.85 (0.80)}          & 0.74 (0.76) & 0.96 (0.93)             & 0.96 (1)             & \textbf{1 (0.98)}             & 0.93 (0.91) \\
4          & 0.78 (0.87)             & 0.78 (0.87)          & \textbf{0.96 (0.91)}          & 0.74 (0.82) & 0.96 (1)                & 0.96 (1)             & \textbf{1 (1)}                & 0.93 (0.93) \\
5          & 0.78 (0.87)             & 0.78 (0.87)          & \textbf{0.96 (0.98)}          & 0.74 (0.87) & 0.96 (1)                & 0.96 (1)             & \textbf{1 (1)}                & 0.93 (0.96) \\ 
\bottomrule
\end{tabular}
\end{footnotesize}
        
    \caption{\label{tb:Results} Bankruptcy prediction results on the test sets: 2019 (2020), for several bag-of-words models: with binary one-hot vectors (Binary), TF-IDF, and mean word-to-vec (W2V) representations, as well as a Longformer classifier, and for single-year vs.~three-year text inputs. 
    Reported metrics are the area-under-the-ROC-curve (AUC), average precision (AP), recall@100, cumul.~accuracy profile ratio (CAP), and cumul.~decile rank (1-5). 
    }
    \label{tb:classification_res}
    \label{main_tab}
\vspace{-2mm}
\end{table*}

\subsection{Evaluation Metrics}\label{sec:metrics}

Following  \citeauthor{mai2019deep} [\citeyear{mai2019deep}], we report the \textbf{Area Under the Receiver Operating Curve} (AUC) as main evaluation metric. 
The AUC is often used to quantify the overall prediction performance of binary decision models. It aggregates the information in the Receiver Operator Curve (ROC), which quantifies the trade-off between the true positive rate (or recall) and the false positive 
rate at various classification thresholds. 
%
However, in certain scenarios, a high true positive rate may be more relevant than a low false positive rate. Therefore, we also report the 
\textbf{Recall@100}. It quantifies the proportion of positive cases (bankrupt firms) present in the 100 highest ranked ones, out of all positive samples (all bankrupt firms in the considered year).
In our context, this metric evaluates the models in their effectiveness to detect as many distressed enterprises as possible for a given budget (e.g., the manpower to investigate a hundred firms). 
The \textbf{Cumulative Accuracy Profile Ratio} (CAP) is a ranking based metric with a strong emphasis on recall of the positive class. 
It summarises the information in the CAP curve, which plots the cumulative proportion of positive samples against the percentage of the ranked data taken into account. 
The \textbf{Cumulative Decile Rank} is also a recall oriented metric. It gives the cumulative proportion of all positive samples (bankrupt firms) in each decile when ranking the samples according to the classifier score.
Although we consider recall more important for the bankruptcy case from the perspective of the `given budget' scenario outlined above, we report a precision oriented metric as well. 
The \textbf{Average Precision} (AP) is the weighted mean of the precision at each classification threshold with the increase in recall as weight. 

\section{Models}
\label{sec:models}
Sections~\ref{sec:ohe}-\ref{sec:w2v} introduce our bag-of-words (BoW) models (which discard word order), followed by a neural sequence encoder model that does account for word order (Section~\ref{sec:lf}), and some training details (Section~\ref{sec:training}).  

\subsection{Binary Bag-of-Words Model}
\label{sec:ohe}
As a trivial baseline (referred to as `Binary') we represent our documents as vocabulary-sized binary vectors with `1' at a particular position indicating the presence of the corresponding word. As vocabulary, all occurring unigrams and bigrams are initially considered as features, and reduced to the 20 most informative ones through univariate feature selection, to be used in a logistic regression classifier. This baseline intends to quantify how well the occurrence of a small set of keywords allows predicting bankruptcy. The model for three-year history is obtained the same way, from the joint BoW over the considered years.



\subsection{TF-IDF Bag-of-Words Model}
\label{sec:tfidf}
The second model is similar to the Binary baseline, but considers \emph{term frequency - inverse document frequency} (TF-IDF) features \cite{manning} rather than binary ones, combined with feature selection and an L2-regularized logistic regression classifier. 
The number of features to retain and the inverse regularisation strength are treated as hyperparameters. 
The three-year model is constructed the same way, after concatenating the texts per year. 

\subsection{Word2Vec Average Embedding Model}
\label{sec:w2v}
As a final bag-of-words model (W2V), we implement the best performing architecture proposed by \citeauthor{mai2019deep} [\citeyear{mai2019deep}], based on the Word2Vec model of  \citeauthor{https://doi.org/10.48550/arxiv.1301.3781} [\citeyear{https://doi.org/10.48550/arxiv.1301.3781}]. First, the pre-processed data is used to train skip-gram word representations of dimension 100 (consistent with \citeauthor{mai2019deep} [\citeyear{mai2019deep}]). 
Documents are then represented by the mean word vector over all occurring words. These serve as input to a two-layer feed-forward neural network with ReLU activations \cite{glorot2011deep} and standard dropout  \cite{srivastava2014dropout}, followed by a sigmoid output. 
During training, we minimize the binary cross entropy loss with an L2-penalty, using the Adam optimizer \cite{kingma2014adam}. 
The learning rate, weight decay (L2-penalty), hidden layer width, and dropout rate 
are treated as hyperparameters. 
When performing classification based on a history of three years, the document representations of each year are concatenated, resulting in a 300-dimensional input to the first hidden layer of the neural network. 

\subsection{Longformer}
\label{sec:lf}
For our most advanced neural model, we encode the documents through the Longformer of \citeauthor{https://doi.org/10.48550/arxiv.2004.05150} [\citeyear{https://doi.org/10.48550/arxiv.2004.05150}]. 
This transformer-based model is able to handle sequences up to 4096 tokens through its attention mechanism that scales linearly with the input text length (as opposed to the quadratic behavior in earlier Transformer models such as BERT \cite{devlin2018bert}). Given the mean document length of over 6k words in our corpus (cf.~Table~\ref{tbl:stats}), we considered the Longformer a plausible baseline.
We process the first 4096 tokens of each document with the Longformer model and retain the 768-dimensional pooled output as the document representation that feeds the same feed-forward classification neural network as described above. For dealing with a history of three years, the individual representations per year are again concatenated, and the input size of the first hidden layer is adjusted accordingly. During training, these representations are kept static (i.e., the Longformer weights are not further fine-tuned on our classification task).

\subsection{Training Details}
\label{sec:training}
The classical models (Sections~\ref{sec:ohe} and~\ref{sec:tfidf}) are implemented in scikit-learn\footnotemark[5] and the hyperparameters are optimised through a grid search procedure. As constructing the vocabulary of all tokens in the training data is expensive, we choose to undersample the majority class until a 90\%-10\% distribution was reached. The neural models (Sections~\ref{sec:w2v} and \ref{sec:lf}) 
are implemented in PyTorch\footnotemark[5] while the Word2Vec model was trained with Gensim\footnotemark[5] and the forward pass through the Longformer was performed with Huggingface\footnotemark[4]. Since hyperparameter optimisation for deep learning models is expensive, we made use of the Tree-Structured Parzen Estimation algorithm to find the optimal hyperparameter settings \cite{bergstra2011algorithms} implemented in Optuna\footnotemark[5].
%
%
The hyperparameters are tuned to maximise the weighted AUC of the 2017 and 2018 validation data, and the obtained values are then used to train the final models using training data up to 2017, to be tested on the 2019 and 2020 test sets.\footnotemark[6]

\footnotetext[5]{Scikit-learn: https://scikit-learn.org/stable/ \\ PyTorch: https://pytorch.org/
\\ Optuna: https://optuna.org/ \\ Gensim= https://radimrehurek.com/gensim/}
\footnotetext[6]{The considered hyperparameter ranges can be accessed through the GitHub repository.}

\section{Results and Discussion}
\label{sec:empres}

Table~\ref{tb:classification_res} presents the out-of-period test performance metrics for our text-based bankruptcy prediction models, taking a single year or three years of history into account. 

When taking a single year of history into account, the W2V model is superior in terms of AUC, recall@100 and CAP while the TF-IDF model achieves the best results in terms of AP. For the 2019 test set, the TF-IDF model contains a slightly higher proportion of positive samples in the first decile but the W2V model is superior from the second decile onwards. When taking three years of history into account, the W2V model achieves the best results for the AUC and CAP metrics while the TF-IDF model performs better with respect to AP and recall@100. When looking at decile rank, the W2V models performs best, having ranked all bankrupt companies in the top 30\% of the samples for the 2019 test set. 

For each model, AUC and CAP are better when taking three years of history into account compared to a single year of history. The same applies for the decile rank (except for the TF-IDF model and the Longformer model in the first decile). AP is generally worse when using a longer history, except for the Binary model with test set 2020 and the Longformer model with test set 2019. The recall@100 metric varies over the two setups.

\begin{table}[]
\centering
\resizebox{\columnwidth}{!}{%
\begin{tabular}{l}
\toprule
\multicolumn{1}{c}{\textbf{Top 15 selected unigrams and bigrams}}\\ 
\midrule
waiver (0.26), \_UNK\_ million (0.21), restructuring (0.21), \\severance (0.20), subordinated (0.20), financial covenant (0.15), \\ indenture (0.14), lender (0.14), interest payment (0.14), \\senior secured (0.14), asset sale (0.12), senior (0.09), \\ cross default (0.09), indebtedness (0.07), \\event default (0.05), credit facility (0.05) \\ \bottomrule
\end{tabular}
}
\caption{\label{fig:words} Top 15 tokens with largest logistic regression coefficients (shown in parentheses) of the Binary bag-of-words model with single year history.}
\vspace{-2mm}
\end{table}



We observe that the Binary models based on a mere 20 keywords perform surprisingly well, although not on par with the TF-IDF and W2V models. Note that the latter are based on many more features (in particular, hyperparameter tuning led for the TF-IDF model to 25.000 (10.000) features for single (three) year history). The relatively good performance of the Binary baseline suggests that the presence of few very informative words is a strong indicator for impending bankruptcy.
As an illustration, we list the top 15 unigrams and bigrams selected by the single year Binary model in table~\ref{fig:words} and underline these features in the extracts in table~\ref{tbl:extract}. 

Furthermore, the Longformer model performs significantly worse than the other models. 
Since we do not finetune the generic pre-trained Longformer model on the our end task, the resulting generic document representations appear unable to capture those features in the text that are important for bankruptcy prediction.

The W2V model leads overall to the best results, in particular for AUC (on which model selection was performed over the validation set) and CAP, and better than the Longformer over the entire line. Even though it is based on the mean representation over all words, it appears the relevant information regarding bankruptcy prediction is still sufficiently present.  As opposed to the Longformer, the W2V document representations come from in-domain data (i.e., pretrained on 10-k reports).

Finally, we critically evaluate the observed performance improvements for the three-year w.r.t. single-year history setting. The Binary and TF-IDF models are by construction unable to distinguish the different years, but in principle the W2V and Longformer models could learn to capture a deteriorating financial situation over three years of history. 
However, when evaluating our final W2V models on the test sets with only complete observations (i.e., discard test instances with missing reports), we get the following results. The single year of history AUC is 0.93 (0.94) and the recall@100 is 0.48 (0.36) while the three year history AUC is 0.93 (0.93) and recall@100 was 0.24 (0.28). These results imply that our models taking three years of history into account only lead to better performance metrics as they are able to generate meaningful predictions for companies with some missing reports. Building more expressive models that can leverage the changes in the documents over the years present an interesting avenue for future research.

\section{Conclusion and Future Work} Bankruptcy prediction models are valuable in many real-world applications and have received considerable research attention. However, assessing actual progress in the field 
is not obvious
due to the lack of a common benchmark. In this work, we introduce such a benchmark for bankruptcy prediction using textual data along with several baseline models that demonstrate the predictive value of the textual data. We give a detailed discussion on our benchmark and evaluation design choices and share our  code to reproduce the experiments. 

In future work, we will focus on more advanced models to take into account the temporal evolution of enterprises' financial situation and more advanced language representations (i.e., by finetuning transformer encoders). We also plan to extend the benchmark with structured financial data to build hybrid prediction models. 



{\footnotesize
\bibliographystyle{style}
\bibliography{main}}

\begin{thebibliography}{}

\bibitem[\protect\citeauthoryear{Altman}{1968}]{altman1968financial}
Edward~I Altman.
\newblock Financial ratios, discriminant analysis and the prediction of
  corporate bankruptcy.
\newblock {\em The journal of finance}, 23(4):589--609, 1968.

\bibitem[\protect\citeauthoryear{Balcaen and Ooghe}{2006}]{balcaen200635}
Sofie Balcaen and Hubert Ooghe.
\newblock 35 years of studies on business failure: an overview of the classic
  statistical methodologies and their related problems.
\newblock {\em The British Accounting Review}, 38(1):63--93, 2006.

\bibitem[\protect\citeauthoryear{Beaver}{1966}]{beaver1966financial}
William~H Beaver.
\newblock Financial ratios as predictors of failure.
\newblock {\em Journal of accounting research}, pages 71--111, 1966.

\bibitem[\protect\citeauthoryear{Beltagy \bgroup \em et al.\egroup
  }{2020}]{https://doi.org/10.48550/arxiv.2004.05150}
Iz~Beltagy, Matthew~E. Peters, and Arman Cohan.
\newblock Longformer: The long-document transformer, 2020.

\bibitem[\protect\citeauthoryear{Bergstra \bgroup \em et al.\egroup
  }{2011}]{bergstra2011algorithms}
James Bergstra, R{\'e}mi Bardenet, Yoshua Bengio, and Bal{\'a}zs K{\'e}gl.
\newblock Algorithms for hyper-parameter optimization.
\newblock {\em Advances in neural information processing systems}, 24, 2011.

\bibitem[\protect\citeauthoryear{Bernanke}{1981}]{bernanke1981bankruptcy}
Ben~S Bernanke.
\newblock Bankruptcy, liquidity, and recession.
\newblock {\em The American Economic Review}, 71(2):155--159, 1981.

\bibitem[\protect\citeauthoryear{Bielecki and
  Rutkowski}{2013}]{bielecki2013credit}
Tomasz~R Bielecki and Marek Rutkowski.
\newblock {\em Credit risk: modeling, valuation and hedging}.
\newblock Springer Science \& Business Media, 2013.

\bibitem[\protect\citeauthoryear{Cecchini \bgroup \em et al.\egroup
  }{2010}]{cecchini2010making}
Mark Cecchini, Haldun Aytug, Gary~J Koehler, and Praveen Pathak.
\newblock Making words work: Using financial text as a predictor of financial
  events.
\newblock {\em Decision Support Systems}, 50(1):164--175, 2010.

\bibitem[\protect\citeauthoryear{Devlin \bgroup \em et al.\egroup
  }{2018}]{devlin2018bert}
Jacob Devlin, Ming-Wei Chang, Kenton Lee, and Kristina Toutanova.
\newblock Bert: Pre-training of deep bidirectional transformers for language
  understanding.
\newblock {\em arXiv preprint arXiv:1810.04805}, 2018.

\bibitem[\protect\citeauthoryear{Glorot \bgroup \em et al.\egroup
  }{2011}]{glorot2011deep}
Xavier Glorot, Antoine Bordes, and Yoshua Bengio.
\newblock Deep sparse rectifier neural networks.
\newblock In {\em Proceedings of the fourteenth international conference on
  artificial intelligence and statistics}, pages 315--323. JMLR Workshop and
  Conference Proceedings, 2011.

\bibitem[\protect\citeauthoryear{Hosaka}{2019}]{HOSAKA2019287}
Tadaaki Hosaka.
\newblock Bankruptcy prediction using imaged financial ratios and convolutional
  neural networks.
\newblock {\em Expert Systems with Applications}, 117:287--299, 2019.

\bibitem[\protect\citeauthoryear{Keasey and Watson}{1987}]{keasey1987non}
Kevin Keasey and Robert Watson.
\newblock Non-financial symptoms and the prediction of small company failure: A
  test of argenti's hypotheses.
\newblock {\em Journal of Business Finance \& Accounting}, 14(3):335--354,
  1987.

\bibitem[\protect\citeauthoryear{Kim and Kang}{2010}]{KIM20103373}
Myoung-Jong Kim and Dae-Ki Kang.
\newblock Ensemble with neural networks for bankruptcy prediction.
\newblock {\em Expert Systems with Applications}, 37(4):3373--3379, 2010.

\bibitem[\protect\citeauthoryear{Kingma and Ba}{2014}]{kingma2014adam}
Diederik~P Kingma and Jimmy Ba.
\newblock Adam: A method for stochastic optimization.
\newblock {\em arXiv preprint arXiv:1412.6980}, 2014.

\bibitem[\protect\citeauthoryear{Loper and Bird}{2002}]{loper2002nltk}
Edward Loper and Steven Bird.
\newblock Nltk: The natural language toolkit.
\newblock {\em arXiv preprint cs/0205028}, 2002.

\bibitem[\protect\citeauthoryear{Loukas \bgroup \em et al.\egroup
  }{2021}]{loukas2021edgarcorpus}
Lefteris Loukas, Manos Fergadiotis, Ion Androutsopoulos, and Prodromos
  Malakasiotis.
\newblock Edgar-corpus: Billions of tokens make the world go round, 2021.

\bibitem[\protect\citeauthoryear{Mai \bgroup \em et al.\egroup
  }{2019}]{mai2019deep}
Feng Mai, Shaonan Tian, Chihoon Lee, and Ling Ma.
\newblock Deep learning models for bankruptcy prediction using textual
  disclosures.
\newblock {\em European journal of operational research}, 274(2):743--758,
  2019.

\bibitem[\protect\citeauthoryear{Manning \bgroup \em et al.\egroup
  }{2008}]{manning}
Christopher~D. Manning, Prabhakar Raghavan, and Hinrich Sch\"{u}tze.
\newblock {\em Introduction to Information Retrieval}.
\newblock Cambridge University Press, USA, 2008.

\bibitem[\protect\citeauthoryear{Matin \bgroup \em et al.\egroup
  }{2019}]{matin2019predicting}
Rastin Matin, Casper Hansen, Christian Hansen, and Pia M{\o}lgaard.
\newblock Predicting distresses using deep learning of text segments in annual
  reports.
\newblock {\em Expert Systems with Applications}, 132:199--208, 2019.

\bibitem[\protect\citeauthoryear{Mayew \bgroup \em et al.\egroup
  }{2015}]{mayew2015md}
William~J Mayew, Mani Sethuraman, and Mohan Venkatachalam.
\newblock Md\&a disclosure and the firm's ability to continue as a going
  concern.
\newblock {\em The Accounting Review}, 90(4):1621--1651, 2015.

\bibitem[\protect\citeauthoryear{Mikolov \bgroup \em et al.\egroup
  }{2013}]{https://doi.org/10.48550/arxiv.1301.3781}
Tomas Mikolov, Kai Chen, Greg Corrado, and Jeffrey Dean.
\newblock Efficient estimation of word representations in vector space, 2013.

\bibitem[\protect\citeauthoryear{Odom and Sharda}{1990}]{odom1990neural}
Marcus~D Odom and Ramesh Sharda.
\newblock A neural network model for bankruptcy prediction.
\newblock In {\em 1990 IJCNN International Joint Conference on neural
  networks}, pages 163--168. IEEE, 1990.

\bibitem[\protect\citeauthoryear{Ohlson}{1980}]{ohlson1980financial}
James~A Ohlson.
\newblock Financial ratios and the probabilistic prediction of bankruptcy.
\newblock {\em Journal of accounting research}, pages 109--131, 1980.

\bibitem[\protect\citeauthoryear{Shumway}{2001}]{shumway2001forecasting}
Tyler Shumway.
\newblock Forecasting bankruptcy more accurately: A simple hazard model.
\newblock {\em The journal of business}, 74(1):101--124, 2001.

\bibitem[\protect\citeauthoryear{Srivastava \bgroup \em et al.\egroup
  }{2014}]{srivastava2014dropout}
Nitish Srivastava, Geoffrey Hinton, Alex Krizhevsky, Ilya Sutskever, and Ruslan
  Salakhutdinov.
\newblock Dropout: a simple way to prevent neural networks from overfitting.
\newblock {\em The journal of machine learning research}, 15(1):1929--1958,
  2014.

\bibitem[\protect\citeauthoryear{Vaswani \bgroup \em et al.\egroup
  }{2017}]{vaswani2017attention}
Ashish Vaswani, Noam Shazeer, Niki Parmar, Jakob Uszkoreit, Llion Jones,
  Aidan~N Gomez, {\L}ukasz Kaiser, and Illia Polosukhin.
\newblock Attention is all you need.
\newblock {\em Advances in neural information processing systems}, 30, 2017.

\bibitem[\protect\citeauthoryear{Veganzones and
  S{\'e}verin}{2018}]{veganzones2018investigation}
David Veganzones and Eric S{\'e}verin.
\newblock An investigation of bankruptcy prediction in imbalanced datasets.
\newblock {\em Decision Support Systems}, 112:111--124, 2018.

\bibitem[\protect\citeauthoryear{Volkov \bgroup \em et al.\egroup
  }{2017}]{volkov2017incorporating}
Andrey Volkov, Dries~F Benoit, and Dirk Van~den Poel.
\newblock Incorporating sequential information in bankruptcy prediction with
  predictors based on markov for discrimination.
\newblock {\em Decision Support Systems}, 98:59--68, 2017.

\bibitem[\protect\citeauthoryear{Wang \bgroup \em et al.\egroup
  }{2020}]{wang2020financial}
Gang Wang, Jingling Ma, Gang Chen, and Ying Yang.
\newblock Financial distress prediction: Regularized sparse-based random
  subspace with er aggregation rule incorporating textual disclosures.
\newblock {\em Applied Soft Computing}, 90:106152, 2020.

\end{thebibliography}

\end{document}